\documentclass[conference]{IEEEtran}
\IEEEoverridecommandlockouts
\usepackage{cite}
\usepackage{amsmath,amssymb,amsfonts}
\usepackage{algorithmic}
\usepackage{graphicx}
\usepackage{textcomp}
\usepackage{xcolor}
\usepackage{booktabs}
\usepackage{multirow}
\usepackage{tabularx}
\usepackage{url}
\usepackage{hyperref}
\usepackage{authblk}
\newcolumntype{Y}{>{\centering\arraybackslash}X}
\newcolumntype{R}{>{\raggedleft\arraybackslash}X}
\usepackage{orcidlink}

\def\BibTeX{{\rm B\kern-.05em{\sc i\kern-.025em b}\kern-.08em
    T\kern-.1667em\lower.7ex\hbox{E}\kern-.125emX}}

\begin{document}

\title{Forestpest-YOLO: A High-Performance Detection Framework for Small Forestry Pests}

\author[2\orcidlink{0009-0003-2385-5343}]{Aoduo Li}
\author[2\orcidlink{0009-0007-9887-7714}]{Peikai Lin}
\author[2]{Jiancheng Li}
\author[1*]{Zhen Zhang\thanks{* Corresponding author.}}
\author[3]{Shiting Wu}
\author[1]{Zexiao Liang}
\author[4]{Zhifa Jiang}

\affil[1]{School of Computer Science and Engineering, Huizhou University}
\affil[2]{School of Advanced Manufacturing, Guangdong University of Technology}
\affil[3]{Huizhou Boluo Power Supply Bureau Guangdong Power Grid Co.,Ltd.}
\affil[4]{Obstetrics and Gynaecology, Huizhou First Maternal and Child Health Care Hospital}

\maketitle

\begin{abstract}
Detecting agricultural pests in complex forestry environments using remote sensing imagery is fundamental for ecological preservation, yet it is severely hampered by practical challenges. Targets are often minuscule, heavily occluded, and visually similar to the cluttered background, causing conventional object detection models to falter due to the loss of fine-grained features and an inability to handle extreme data imbalance. To overcome these obstacles, this paper introduces Forestpest-YOLO, a detection framework meticulously optimized for the nuances of forestry remote sensing. Building upon the YOLOv8 architecture, our framework introduces a synergistic trio of innovations. We first integrate a lossless downsampling module, SPD-Conv, to ensure that critical high-resolution details of small targets are preserved throughout the network. This is complemented by a novel cross-stage feature fusion block, CSPOK, which dynamically enhances multi-scale feature representation while suppressing background noise. Finally, we employ VarifocalLoss to refine the training objective, compelling the model to focus on high-quality and hard-to-classify samples. Extensive experiments on our challenging, self-constructed ForestPest dataset demonstrate that Forestpest-YOLO achieves state-of-the-art performance, showing marked improvements in detecting small, occluded pests and significantly outperforming established baseline models.
\end{abstract}


\section{Introduction}
Automated remote sensing monitoring of forestry pests is a key technology for achieving smart forestry and preventing ecological disasters. High-resolution images acquired using unmanned aerial vehicles (UAVs) enable object detection to identify and locate pest outbreak areas with efficiency and coverage far beyond manual inspection. However, this task remains challenging.

A primary difficulty lies in the extreme scales and low signal-to-noise ratio of targets. Pests such as individual eggs or boreholes are extremely small; in our dataset, over 70\% of targets have an area less than $32 \times 32$ pixels. These weak cues are easily overwhelmed by high-frequency noise such as bark textures and dappled shadows. Severe occlusion and complex forest backgrounds compound the issue: canopy structure means targets are often occluded by leaves and branches. Backgrounds also contain non-target objects (rocks, fallen leaves) similar in color/shape to pest camouflage, leading to high false detections. Finally, data acquisition and generalization present significant hurdles. Collection varies by season and weather, and pest morphology varies across life-cycle stages, increasing intra-class variance. Existing datasets often lack negative images, causing many false positives in healthy areas and limiting generalization. Similar concerns about imbalance and generalization appear in semi-supervised medical image segmentation and contrastive learning \cite{li2025gre,zheng2024lagrange}.

To address these challenges, we introduce Forestpest-YOLO, a multi-pronged detector built on YOLOv8 \cite{yolov8}. First, to avoid feature loss on minute targets, we integrate SPD-Conv \cite{sapan2022spd} to preserve fine detail. Second, we propose the CSPOK module, leveraging Omni-Kernel \cite{li2024omnikernel} for robust, efficient multi-scale fusion. Finally, we employ VarifocalLoss \cite{zhang2021varifocalnet} to handle severe sample imbalance. Given the visual complexity and occlusion in natural scenes, advances in robust visual representation—especially secure, noise-resilient medical image encryption using permutation–diffusion and chaotic dynamics—offer useful design insights \cite{le2024medical,xu2024plaintext,li2025dppad,zhong2025image}. We validate efficacy via extensive experiments on our custom ForestPest dataset.

\section{Related Work}
Our research sits at the intersection of three areas in object detection and computer vision \cite{chen1,chen2,chen3,chen4,chen5,chen6,chen7,chen8,zhang1,zhang2,zhang3,zhang4,zhang5,zhang6,zhang7,zhang8,zhang9,zhang10,zhang11,zhang12,zhang13,zhang14,zhang15}: small object detection \cite{wang2023qgd,wang2024beyond,wang2025structure,wang2025visioncube,liu2019convolutional,liu2020fine,zhu2024test,li2024cross,liu2024forgeryttt,zhang2022correction}, feature fusion and attention \cite{li2022monocular,li2023cee,li2022few,li2025adaptive,li2021hybrid,liu2023explicit,liu2023coordfill,zheng2024smaformer,liu2024dh,jiang2020geometry,liu2024depth}, and loss design for robust training \cite{cai2024relation,chen2024adaptive,wu2024image,wu2025llm,wuimgfu,wu2025prompt,chen9,chen10,chen11,chen12,chen13,chen14,chen15}.

\subsection{Advances in Small Object Detection}
Small object detection remains difficult due to scarce distinguishable features \cite{tong2020recent}. A foundational route is multi-scale representation, popularized by FPN \cite{lin2017feature}, which fuses high-level semantics with low-level detail. Yet aggressive backbone downsampling can degrade fine information before fusion. Lossless alternatives like SPD-Conv \cite{sapan2022spd} fold spatial information into channel depth, central to our design. PANet \cite{panet} adds a bottom-up path, and BiFPN \cite{bifpn_efficientdet} introduces weighted, bi-directional fusion. Data-centric strategies—copy-paste \cite{copy_paste_aug}, Mosaic, MixUp—enrich tiny-instance exposure; GAN-based super-resolution \cite{gan_sod} and context modeling \cite{context_sod} further help.

\subsection{Feature Fusion and Attention Mechanisms}
Attention improves fusion across scales. SE-Net \cite{hu2018squeeze} introduced channel attention; CBAM \cite{woo2018cbam} added spatial attention. To capture global context, Non-local Networks \cite{non_local} model long-range dependencies; ViT \cite{vit} and DETR \cite{detr} rely on full self-attention but at notable cost. Dynamic, content-aware convolutions—Deformable Convolutions \cite{deformable_conv} and Dynamic Convolutions \cite{dynamic_conv}—adapt sampling or weights. Our CSPOK follows this trend, leveraging Omni-Kernel \cite{li2024omnikernel} for dynamic, efficient fusion tailored to forestry-pest textures and structures.

\subsection{Loss Functions and Training Strategies for Robust Training}
Loss design is critical for imbalanced, hard samples. Focal Loss \cite{lin2017focal} down-weights easy examples; VarifocalLoss (VFL) \cite{zhang2021varifocalnet} treats positives/negatives asymmetrically, prioritizing high-quality positives. For localization, Smooth L1 has given way to IoU-based losses (GIoU/DIoU/CIoU \cite{giou,ciou}) that better match evaluation. GFL \cite{gfl} learns box distributions, and dynamic assignment (ATSS \cite{atss}, SimOTA \cite{yolox}) improves stability. These advances yield more stable, effective training, especially on challenging sets like ForestPest.

\section{Methodology: The Forestpest-YOLO Framework}
The architectural design of Forestpest-YOLO is centered on three strategic modifications to the YOLOv8 framework, each targeting a specific challenge in forestry pest detection. We engineered a novel framework that not only preserves critical low-level features but also enhances multi-scale feature interaction and optimizes the learning objective for imbalanced data.

\subsection{Overall Architecture}
\begin{figure}[htbp]
\centering
\includegraphics[width=\linewidth]{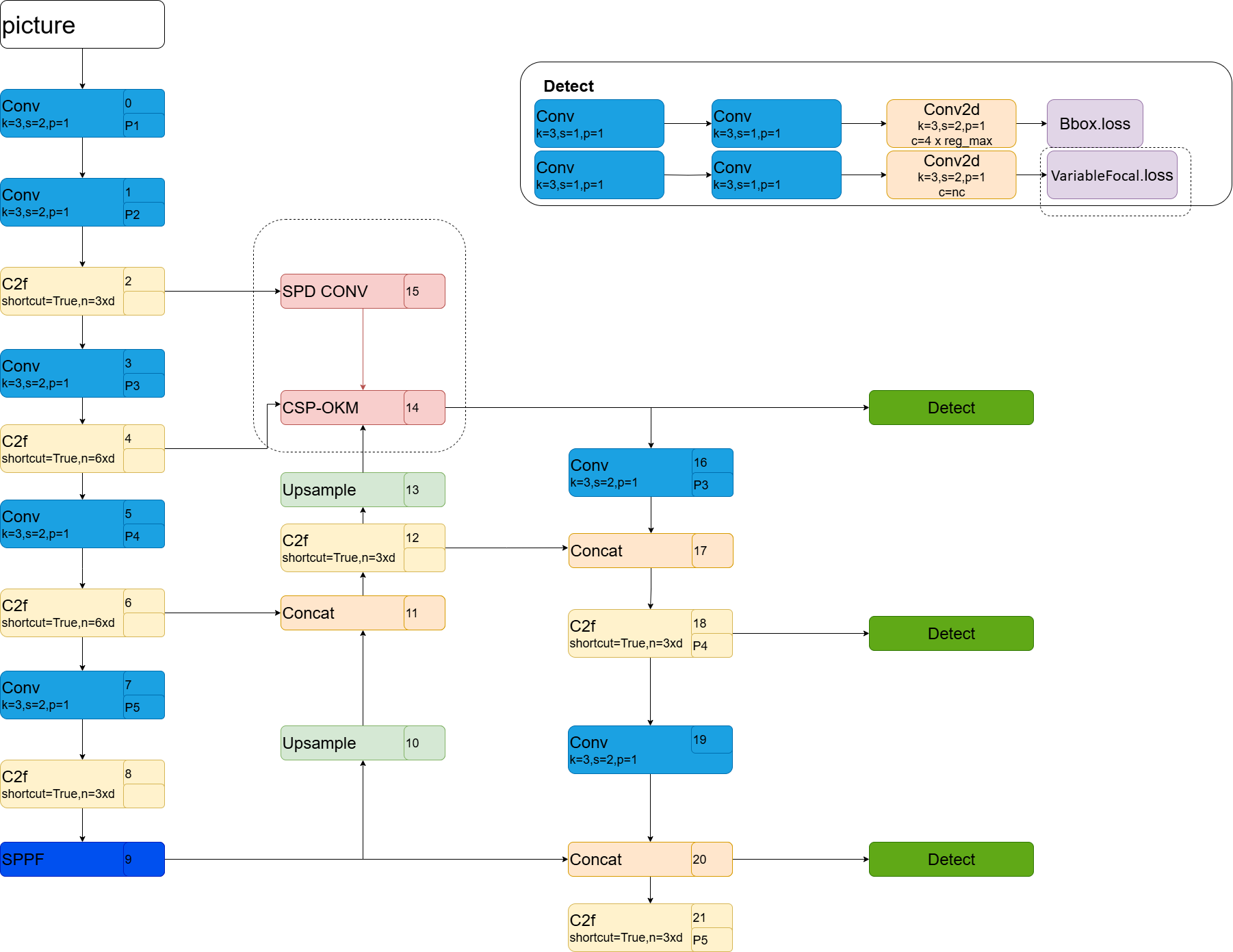}
\caption{The overall network architecture of Forestpest-YOLO, illustrating the main data flow from input to detection. Key modifications include the integration of the SPD-Conv module in the backbone and the CSPOK fusion block in the neck.}
\label{fig:overall_architecture}
\end{figure}

The overall architecture of Forestpest-YOLO, illustrated in Fig. \ref{fig:overall_architecture}, is a systematic enhancement of the standard YOLOv8 pipeline. The data flow begins in the backbone, where we strategically replace a conventional strided convolution with our SPD-Conv module after the P2 feature extraction stage. This crucial intervention ensures that high-resolution spatial information, which is vital for small object detection, is preserved rather than discarded during downsampling. The feature maps then proceed to the neck, where we substitute the standard C2f fusion units with our more powerful CSPOK modules. These modules are designed to facilitate a more sophisticated and robust integration of features across different semantic and spatial scales. Finally, the fused feature maps are passed to the detection heads, where the learning process is governed by VarifocalLoss, replacing the original classification loss to better cope with the severe class imbalance typical in pest detection scenarios.

\subsection{SPD-Conv: Lossless Downsampling Module}
To address the degradation of fine-grained features caused by traditional downsampling, we incorporate SPD-Conv, a space-to-depth feature transformer. Given an input feature map $X \in \mathbb{R}^{S \times S \times C_1}$, SPD-Conv first performs a slicing operation with a scale factor (e.g., scale=2). This partitions $X$ into scale$^2$ sub-maps. For scale=2, the four sub-maps are defined by:
\begin{equation}
\begin{aligned}
f_{0,0} = X[0::2, 0::2], \quad f_{0,1} = X[0::2, 1::2] \\
f_{1,0} = X[1::2, 0::2], \quad f_{1,1} = X[1::2, 1::2]
\end{aligned}
\end{equation}
These sub-maps are then concatenated along the channel dimension, transforming the spatial information into channel depth and creating an intermediate feature map $X'$:
\begin{equation}
X' = \text{Concat}(f_{0,0}, f_{0,1}, f_{1,0}, f_{1,1})
\end{equation}
where $X' \in \mathbb{R}^{\frac{S}{2} \times \frac{S}{2} \times 4C_1}$. Finally, a non-strided convolution is applied to reduce the channel dimension and learn richer feature representations, producing the output $X''$:
\begin{equation}
X'' = \text{Conv}_{1 \times 1}(X')
\end{equation}
where $X'' \in \mathbb{R}^{\frac{S}{2} \times \frac{S}{2} \times C_2}$. This process effectively halves the spatial dimensions while preserving the complete feature set.

\subsection{CSPOK: Cross-Stage Parallel Omni-Kernel Fusion}
\begin{figure}[htbp]
\centering
\includegraphics[width=\linewidth]{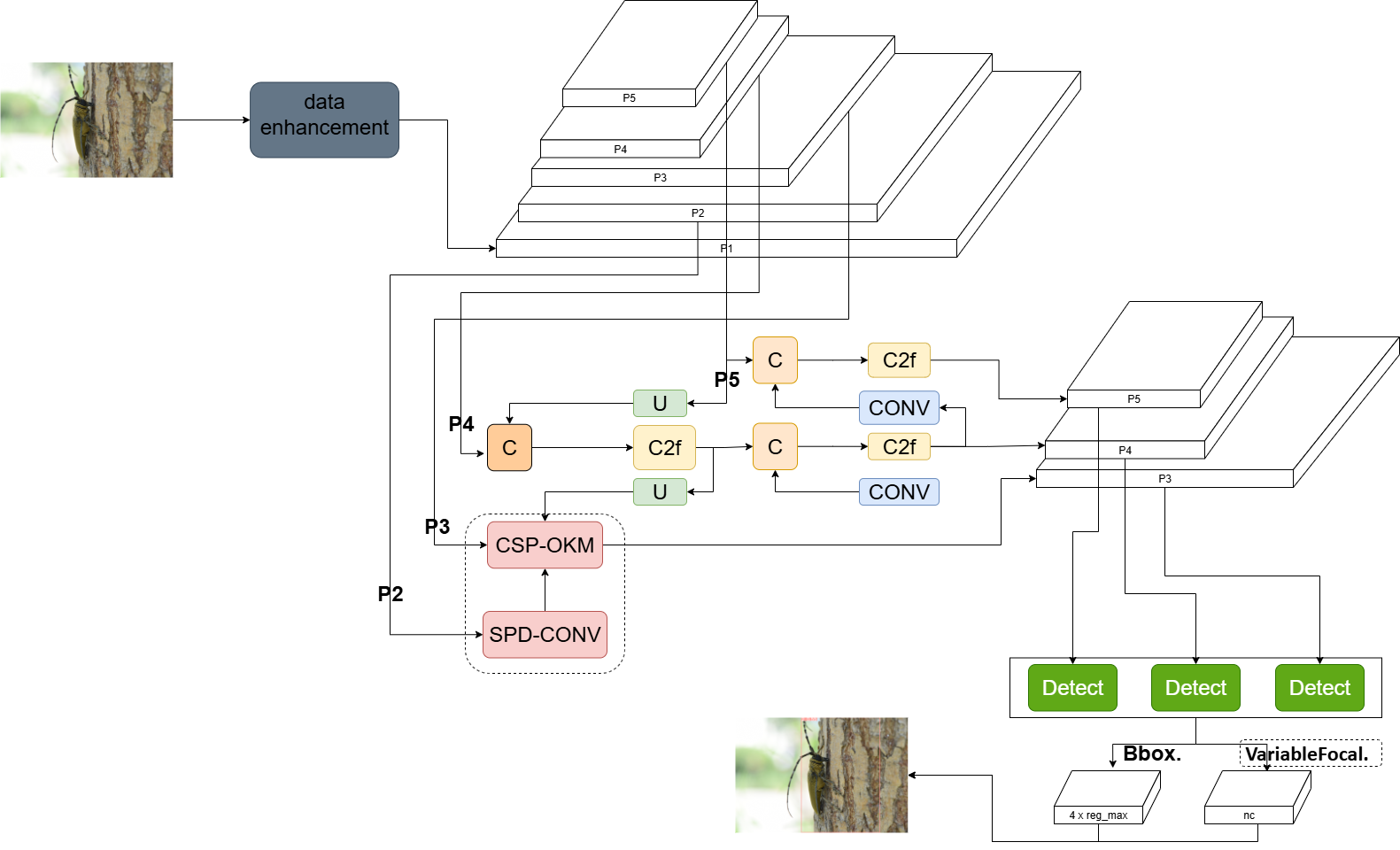}
\caption{Detailed structure of the improved neck, highlighting the integration of our proposed modules for enhanced feature fusion. The diagram shows SPD-Conv for lossless downsampling and the CSPOK block replacing the original C2f unit.}
\label{fig:neck_architecture}
\end{figure}

The CSPOK module, whose structure is detailed in Fig. \ref{fig:neck_architecture}, is designed for superior multi-scale feature fusion by combining the efficiency of Cross-Stage Partial (CSP) design with the adaptability of Omni-Kernel (OKM). An input feature map $F_{in}$ is first split into two parts:
\begin{equation}
F_{in} \rightarrow [F_{1}, F_{2}]
\end{equation}
$F_1$ serves as a direct, cross-stage connection, preserving the original information flow. The other part, $F_2$, is passed through a sophisticated processing block where the Omni-Kernel mechanism is applied. This is represented as:
\begin{equation}
F'_{2} = \text{OKM}(\text{Conv}(F_{2}))
\end{equation}
The OKM block dynamically adapts its fusion strategy based on the input features. The two pathways are then concatenated and passed through a final convolutional layer to integrate the information:
\begin{equation}
F_{out} = \text{Conv}(\text{Concat}(F_{1}, F'_{2}))
\end{equation}
This parallel design enriches the feature diversity and enhances the model's ability to capture both local details and global context.

\subsection{VarifocalLoss (VFL) and Matching Optimization}
\begin{figure}[htbp]
\centering
\includegraphics[width=\linewidth]{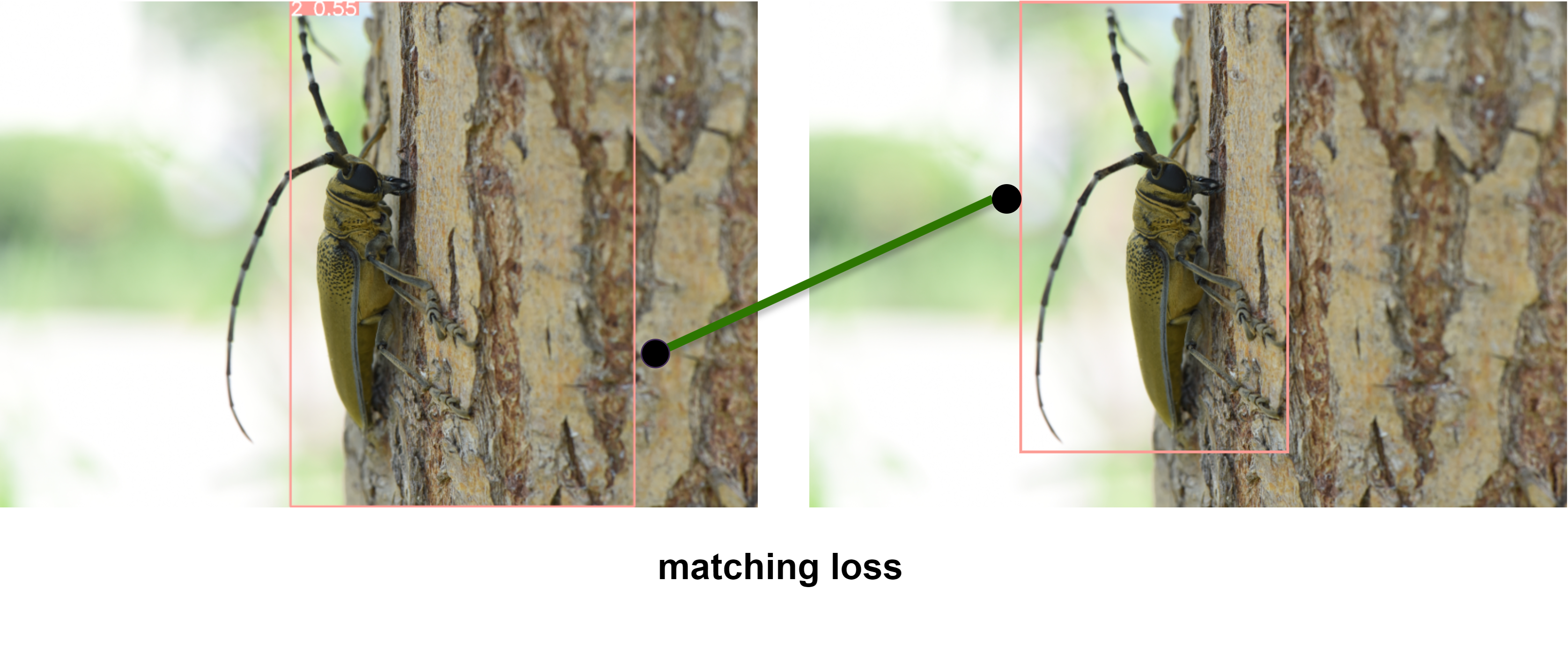}
\caption{Conceptual visualization of matching quality improvement achieved by using VarifocalLoss. The left side depicts potential poor matching, while the right (green lines) illustrates the precise target association our loss encourages.}
\label{fig:matching_loss}
\end{figure}

To address the severe imbalance between easy/hard and positive/negative samples, we adopt VarifocalLoss (VFL) \cite{zhang2021varifocalnet} as the classification loss function:
\begin{equation}
L_{VFL}(p,q) =
\begin{cases}
-q(q\log(p) + (1-q)\log(1-p)) & q > 0, \\
-\alpha p^\gamma \log(1-p) & q=0,
\end{cases}
\label{eq:vfl}
\end{equation}
where $p$ is the model's predicted score, and $q$ is the target IoU score. Based on empirical validation, we set the hyperparameters to $\alpha=0.75$ and $\gamma=2.0$. As conceptually illustrated in Fig. \ref{fig:matching_loss}, VFL's asymmetric weighting scheme encourages the model to focus on high-quality positive samples, leading to more precise localization and classification.

\begin{table*}[ht]
\centering
\caption{Performance Comparison on the ForestPest Test Set}
\label{tab:main_results}
\begin{tabularx}{\textwidth}{@{}l*{5}{Y}R@{}}
\toprule
\textbf{Model} & \textbf{mAP\textsuperscript{.5:.95}} & \textbf{mAP\textsuperscript{.5}} & \textbf{AP\textsuperscript{small}} & \textbf{Params(M)} & \textbf{FLOPs(G)} & \textbf{FPS} \\
\midrule
YOLOv5s & 0.478 & 0.728 & 0.110 & 7.2 & 16.5 & 140 \\
YOLOv8s & 0.482 & 0.746 & 0.112 & 11.2 & 28.6 & 125 \\
RT-DETR-R50 & 0.501 & 0.737 & 0.128 & 33.0 & 109.0 & 74 \\
\midrule
\textbf{Forestpest-YOLO (Ours)} & \textbf{0.508} & \textbf{0.762} & \textbf{0.131} & \textbf{12.1} & \textbf{30.2} & \textbf{118} \\
\bottomrule
\end{tabularx}
\end{table*}

\section{Experiments}
\subsection{Dataset: The ForestPest Dataset}
To effectively evaluate our proposed model, we constructed a self-built forestry remote sensing pest detection dataset named \textbf{ForestPest}. This dataset is designed to simulate various challenges encountered in real-world forestry monitoring, containing 5,690 high-resolution UAV images of common forestry pests in COCO format. The dataset is characterized by its diversity in scenes, pest species, and life stages, with a significant portion of small, occluded, and camouflaged targets, making it a challenging benchmark. Key statistics are summarized in Table \ref{tab:dataset_stats}.

\begin{table}[htbp]
\centering
\caption{ForestPest Dataset Statistics}
\label{tab:dataset_stats}
\begin{tabularx}{\columnwidth}{@{}lX@{}}
\toprule
\textbf{Attribute} & \textbf{Value} \\
\midrule
Total Images & 5,690 \\
Train/Val Split & 4,800 / 890 \\
Pest Classes & 15 \\
Total Bounding Boxes & 32,450 \\
Small Targets ($<32^2$ pixels) Ratio & bigger than 70\% \\
\bottomrule
\end{tabularx}
\end{table}

\subsection{Implementation Details}
All models were trained under a unified experimental protocol to ensure a fair comparison. The implementation was based on PyTorch, with YOLOv8s serving as the baseline. The experiments were conducted on a server equipped with four NVIDIA RTX 3090 GPUs. We employed the AdamW optimizer with an initial learning rate of $1 \times 10^{-3}$, which was adjusted using a cosine annealing strategy over 100 training epochs. Input images were uniformly resized to 640x640, and a batch size of 8 was used. To enhance model robustness and prevent overfitting, we applied a suite of data augmentation techniques, including Mosaic, MixUp, random affine transformations (rotation, scaling), and color jitter. The $AP_{small}$ metric is calculated following the COCO standard, defining small objects as those with an area less than $32^2$ pixels.

\subsection{Comparison with State-of-the-Art Models}
We conducted a comprehensive performance comparison between Forestpest-YOLO and several mainstream object detectors on the ForestPest test set, with results shown in Table \ref{tab:main_results}. Our model demonstrates superior performance across all key accuracy metrics. Notably, Forestpest-YOLO achieves an mAP@.5:.95 of 0.508, surpassing the YOLOv8s baseline. The most significant improvement is observed in the $AP_{small}$ metric, where our model achieves 0.131, a 17.0\% relative increase over YOLOv8s. This highlights the efficacy of our framework in addressing the core challenge of small object detection in complex forestry environments. While other models like YOLOv8s show competitive precision, their lower recall suggests a tendency to miss difficult targets, a shortcoming our model effectively addresses. These accuracy gains are achieved with only a marginal increase in computational cost, making it a practical solution for real-world applications.

\begin{table}[ht]
\centering
\caption{Ablation Study of Forestpest-YOLO Components}
\label{tab:ablation}
\begin{tabularx}{\columnwidth}{@{}lYYY@{}}
\toprule
\textbf{Configuration} & \textbf{mAP\textsuperscript{.5:.95}} & \textbf{mAP\textsuperscript{.5}} & \textbf{AP\textsuperscript{small}} \\
\midrule
YOLOv8s (Baseline) & 0.482 & 0.746 & 0.112 \\
+ SPD-Conv & 0.491 & 0.750 & 0.119 \\
+ SPD-Conv + CSPOK & 0.499 & 0.753 & 0.126 \\
\textbf{+ VFL (Full Model)} & \textbf{0.508} & \textbf{0.762} & \textbf{0.131} \\
\bottomrule
\end{tabularx}
\end{table}

\subsection{Ablation Study}
To dissect the individual contributions of our proposed components, we performed a systematic ablation study. We began with the YOLOv8s baseline and incrementally added each module: SPD-Conv, CSPOK, and finally VarifocalLoss (VFL). The results, presented in Table \ref{tab:ablation}, clearly demonstrate the effectiveness of each enhancement. The introduction of SPD-Conv alone yielded a 6.3\% relative improvement in $AP_{small}$, confirming the benefits of its lossless downsampling for small object feature preservation. Building on this, the addition of the CSPOK module further boosted performance, validating its superior multi-scale fusion capabilities. The final integration of VFL provided an additional lift across all metrics, culminating in our full Forestpest-YOLO model, which achieved the highest scores. This step-by-step analysis validates that our modifications work synergistically.

\section{Discussion}
\subsection{Analysis of Method Effectiveness}
Our method's success stems from its targeted design for forestry scenarios. SPD-Conv preserves critical high-frequency details. CSPOK simultaneously processes fine-grained local features and broader context. VarifocalLoss ensures training focuses on informative and challenging samples rather than being dominated by easy background examples.

\subsection{Limitations and Future Work}
Despite the significant progress, our model has limitations. Performance may decline under extreme weather. Its reliance on supervised learning requires large annotated datasets. Future work will explore multimodal data fusion, semi-supervised learning, and integrating lightweight Transformer-based decoders.

\section{Conclusion}
This paper introduced Forestpest-YOLO, a framework specifically engineered to overcome the critical challenges in forestry pest detection. By synergistically integrating SPD-Conv, the CSPOK module, and VarifocalLoss, our approach achieves a new state-of-the-art on the challenging ForestPest benchmark. These advancements not only provide a powerful and practical tool for solving real-world problems in remote sensing but also offer valuable insights for designing future intelligent and robust visual detection systems.

\section*{Acknowledgment}
This research was funded by the Young Innovative Talents Project of Colleges and Universities in Guangdong Province (2021KQNCX092); the Doctoral Program of Huizhou University (2020JB028); and the Outstanding Youth Cultivation Project of Huizhou University (HZU202009).

\bibliographystyle{IEEEtran}
\bibliography{ref}

\end{document}